# Investigating image-based fallow weed detection performance on *Raphanus sativus* and *Avena sativa* at speeds up to 30 km h$^{-1}$.

Guy R. Y. Coleman[a], Angus Macintyre[a], Michael J. Walsh[a], William T. Salter[b]

*Guy Coleman ([guy.coleman@sydney.edu.au](guy.coleman@sydney.edu.au)) is the corresponding author*

[a] *School of Life and Environmental Sciences, Sydney Institute of Agriculture, The University of Sydney, Brownlow Hill, NSW, Australia*

[b] *School of Life and Environmental Sciences, Sydney Institute of Agriculture, The University of Sydney, Narrabri, NSW, Australia*

## 1 Abstract

Site-specific weed control (SSWC) can provide considerable reductions in weed control costs and herbicide usage. Despite the promise of machine vision for SSWC systems and the importance of ground speed in weed control efficacy, there has been little investigation of the role of ground speed and camera characteristics on weed detection performance. Here, we compare the performance of four camera/software combinations using the open-source OpenWeedLocator platform – (1) default settings on a Raspberry Pi HQ camera; (2) optimised software settings on a HQ camera, (3) optimised software settings on the Raspberry Pi v2 camera; and (4) a global shutter Arducam AR0234 camera – at speeds ranging from 5 km h-1 to 30 km h-1. A combined excess green (ExG) and hue, saturation, value (HSV) thresholding algorithm was used for testing under fallow conditions using tillage radish (*Raphanus sativus*) and forage oats (*Avena sativa*) as representative broadleaf and grass weeds, respectively. ARD demonstrated the highest recall among camera systems, with up to 95.7% of weeds detected at 5 km h$^{-1}$ and 85.7% at 30 km h$^{-1}$. HQ1 and V2 cameras had the lowest recall of 31.1% and 26.0% at 30 km h$^{-1}$, respectively. All cameras experienced a decrease in recall as speed increased. The highest rate of decrease was observed for HQ1 with 1.12% and 0.90% reductions in recall for every km h$^{-1}$ increase in speed for tillage radish and forage oats, respectively. Detection of the 'grassy' forage oats was worse (P<0.05) than the broadleaved tillage radish for all cameras. Despite the variations in recall, HQ1, HQ2, and V2 maintained near-perfect precision at all tested speeds. The variable effect of ground speed and camera system on detection performance of grass and broadleaf weeds, indicates that careful hardware and software considerations must be made when developing SSWC systems.

*Keywords*: site-specific weed control, machine vision, precision agriculture, Raspberry Pi, open-source

## 2 Introduction

The practice of targeting weeds and weed patches individually is known as site-specific weed control (SSWC), which offers significant opportunities for more efficient and effective weed management (Lopez-Granados, 2011; Mortensen et al., 1995). SSWC transforms the weed control-cost equation from one based on area to one based on weed density. It is an attractive prospect for growers, with substantial savings in herbicide inputs (Timmermann et al., 2003) and/or can reduce the area of tillage required (Walsh et al., 2020) when weed densities are low. Importantly, it is also an opportunity for the viable deployment of alternative weed control options (e.g., lasers, waterjet cutting and abrasive grit) that would be impractical in large-scale production systems (Coleman et al., 2019). Yet, as identified over many years (Thompson et al., 1991), the major bottleneck for SSWC is in implementing a reliable form of weed detection that provides (1) appropriate performance; (2) class specificity ('weed' and 'crop', or species level) and (3) detection granularity (classification, object detection or segmentation) at the desired speed, in variable conditions and within a specific weed control scenario (e.g., fallow or in-crop).

In a fallow weed control scenario, weed detection enables the site-specific control of weeds for soil water and nutrient conservation in rainfall limited cropping systems (Thomas et al., 2007; Verburg et al., 2012). Current SSWC in fallow is enabled by the principle that all green, actively growing plants in a field are weeds. This allows the use of simple photoelectric systems for weed detection such as the WeedIT and WeedSeeker (Peteinatos et al., 2014). The approach generally relies on differential reflection in red and near-infrared spectra between plants and the background (soil and/or crop residues) (Palmer and Owen, 1971). The practice is common in some Australian cropping regions, where reflectance-based systems have been widely adopted (SPAA, 2016). Such systems have been available for commercial use since the late 90s (Shearer and Jones, 1991; Visser and Timmermans, 1996) and have allowed for relatively high performing weed detection (approximately 90%) at speeds up to 20 km h$^{-1}$ (Timmermann et al., 2003).

Suitable environmental conditions for the safe and effective application of herbicides are frequently limited, thus from a farmer perspective, ground speed is a critical factor in the timely delivery of weed control treatments. For example, in Australia's summer cropping region, the combination of high temperatures, humidity and inversion layers frequently limit the time available for herbicide application. This time limitation is a considerable incentive for adoption of higher ground speeds for herbicide application (Butts et al., 2021) and larger spray equipment sizes, however, associated increases in equipment cost, weight and breakdowns are substantial limitations. Additionally, higher ground speed during application limits the efficacy of herbicide delivery by influencing droplet characteristics and spray patterns as well as increasing the potential for spray drift (Meyer et al., 2016; Wolf et al., 1997). In large-scale grain production systems, the limited available time to cover the production area necessitates a careful balance between ground speed and herbicide efficacy. Importantly, in a SSWC system, ground speed plays a role in influencing weed detection performance.

While reflectance-based weed detection systems set a benchmark for performance and encouraged interest in the use of SSWC systems, contemporary SSWC research has moved towards the use of computer vision for weed detection (Coleman et al., 2022a; Wang et al., 2019). Initial attempts at real-time weed detection were largely restricted to speeds under 1 km h$^{-1}$ (Åstrand and Baerveldt, 2002; Lee et al., 1999). These systems used large, slow computers with low resolution cameras. More recently, small form factor single board computers (SBCs) with relatively high computational power have been employed for real-time weed detection. For example, Calvert et al. (2021) used an NVIDIA Jetson with embedded GPU to run a MobileNetv2 deep learning model for real-time detection and control of Harrisia cactus (*Harrisia martinii*) at approximately 7 km h$^{-1}$. Chechliński et al. (2019) employed the more resource constrained Raspberry Pi to deploy a custom U-Net and Mobilenet architecture at a ground speed of 4.6 km h$^{-1}$. Current commercial systems report effective performance up to 15 km h$^{-1}$ (Martin, 2021), however, are still in their infancy and are only suitable for certain applications. Demonstrating high speed image capture potential, image data has been collected on all-terrain vehicles at up to 50 km h$^{-1}$, although in this case the intent was simply for data collection without deployment of computationally intensive weed detection algorithms (Laursen et al., 2017). In our initial research on open-source, colour-based weed recognition for the detection of weeds in fallow, ground speed was limited to a walking pace of approximately 4 km h$^{-1}$ (Coleman et al., 2022b). The approach lacked an analysis of speed and was limited by the default camera settings of the Raspberry Pi HQ camera. Whilst computational speed is a primary consideration, other computational, hardware and agronomical factors also limit maximum ground speed.

Bringing together the sequence of events that determine SSWC performance (both speed and quality), we propose an event timeline for site-specific herbicide application. It sets a framework for system events that must be completed to successfully target and control a weed (Figure 1), incorporating three stages of (1) detection, (2) actuation and (3) delivery. The time required for completion of this event timeline determines the maximum ground speed for the weed control operation. From the moment a weed enters the field of view of the sensor, the weed detection stage begins (e1.1 – e1.3 in Figure 1). The timing of this step is largely dependent on the camera system, the efficiency of the software and the processing speed of the hardware. Following detection, actuation speed of the relay or transistor after receiving the activation signal combined with the delay in activation of the solenoid (e2.1 – e2.2 in Figure 1). Finally, the movement of herbicide from behind the solenoid through the nozzle, across the gap above the plant and onto the target is the final source of delay (e3.1 – e3.2 in Figure 1). This framework contextualises opportunities for improved operational speed in SSWC systems.

Despite the importance of vehicle forward speed in application efficacy and timeliness, there is limited quantitative analysis on the impact of ground speed on the performance of weed detection hardware and software for e0 – e1.3. In one of the few efforts at measuring the effects of speed on weed detection, Steward et al. (2002) reported weed detection accuracy dropped from 96% at 3.2 – 3.9 km h$^{-1}$ to 86% at 11 – 14 km h$^{-1}$; though, processor and camera performance has advanced substantially in the 20 years since this research. Liu et al. (2021) found the recall of three different deep learning architectures, AlexNet, GoogleNet and VGG-16, all decreased by 9% when increasing speed from 1 km h$^{-1}$ to 5km h$^{-1}$. Moreover, understanding the influence of weed morphotypes (e.g., grass or broadleaf) and increasing ground speed on detection accuracy is important in ensuring high performance image-based weed recognition in all detection scenarios. Processing speed is generally reported in weed recognition research as a measure of algorithm relevance for real-time use in field settings (Hasan et al., 2021). Yet, many other factors such as image blur, shutter speed, dust and wind associated with high-speed movement over a field for weed control application will influence performance. Unfortunately, these practicalities are rarely incorporated into weed recognition research.

To better understand the impacts of ground speed on camera-based weed detection from stages e0 – e1.3, our study employed the OpenWeedLocator (OWL) – a low-cost open source weed detection unit that was developed for research and community use (Coleman et al., 2022b). We modified the original OWL system to run comparisons of four camera systems. The aims of this study were to i) practically assess camera and algorithm performance for fallow weed detection as influenced by incremental increases in speeds from 5 km h$^{-1}$ to 30 km h$^{-1}$ and ii) assess the influence of weed type, (grass or broadleaf) and growth stage on camera system performance.

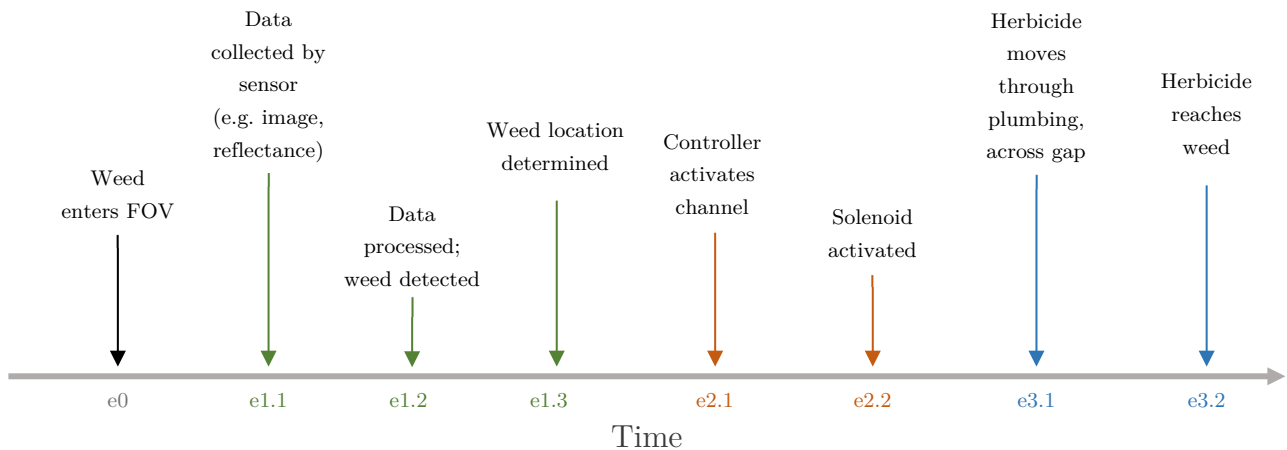

**Figure 1** Overview of key events that must occur from the point a weed enters the field of view of a camera before the herbicide is applied to the plant. The overall time available is determined by the distance between camera and application point and ground speed. The duration of each event is influenced by software and hardware design limitations. Algorithm speed evaluations are typically constrained to e1.2. The present study incorporates stages e0 – e1.3.

## 3 Materials and Methods

### 3.1 Field preparation

Six 25 x 1 m test transects were established at the University of Sydney Lansdowne Farm, Cobbitty, NSW, Australia (-34.022115, 150.664842) in April 2022. Weeds were controlled before seeds of forage oats (Avena sativa) and tillage radish (Raphanus sativus) were sown in a random pattern across each transect as representative broadleaf and grass 'weeds'. Seed was hand planted to a depth of approximately 2 cm on a weekly basis for five weeks to establish plants of variable growth stage and size at a target density of approximately 3 plants m$^{-2}$. To maintain a simulated fallow environment any post germination of tillage radish, forage oats, non-target weed species and moss were selectively spot-treated with non-selective herbicides or removed manually.

### 3.2 Data collection system

A custom vehicle-mounted system was developed for simultaneous video data collection from multiple cameras (Figure 2). All mounts and camera housing components were designed using Tinkercad design software (Autodesk, San Francisco CA, 2012) and produced using a 3D printer (i3 MK3; Prusa Research, Prague, Czech Republic). Four camera systems were used (Table 1); V2, HQ1, HQ2 and ARD. Cameras were mounted in line with each other, within a 3D printed housing to allow for consistency of each field of view (FOV) and synchronisation of data collection (Figure 2). Cameras were positioned at a height of 1 m above the ground, and the FOV was checked such that all weeds within the 1 m wide transects were visible. Each camera was connected to a Raspberry Pi 4 8GB (Raspberry Pi Foundation, Cambridge, UK) embedded computer to run the OWL detection software and record video. Power was supplied by a 12 V battery located in the rear of the vehicle and converted to 5V for each Raspberry Pi using a Pololu (Pololu Corporation, Las Vegas, Nevada) D24V50F5 5V, 5A step-down voltage regulator. Recording for each camera/Raspberry Pi combination was manually turned on and off at the start and end of each transect using a switch connected to the GPIO pins of each Raspberry Pi. A 12 V, 90 W LED work light (Stedi, Melbourne, Victoria, Australia) provided additional lighting within the camera FOV.

All software used for data collection is available from the OpenWeedLocator GitHub repository: https://github.com/geezacoleman/OpenWeedLocator. Specific camera details and settings that were modified from default are included in Table 1.

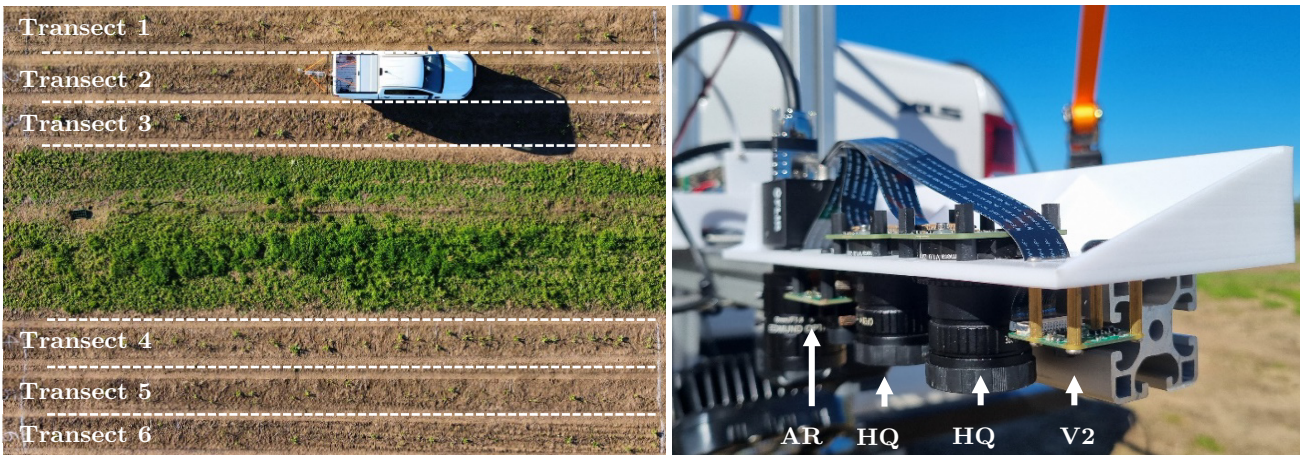

**Figure 2** Aerial view of the field site showing the six, 25 × 1 m transects (left) and the four cameras mounted to the data collection system on the back of the test vehicle (right). Data for each camera were collected simultaneously at each speed and transect to ensure consistency in environmental conditions such as lighting and speed. The vehicle was run at set speeds using a speedometer and GPS-based speed checker over each transect whilst cameras were recording.

**Table 1** Camera and software configurations used to evaluate the effect of increasing speed on camera and weed detection algorithm performance for tillage radish (*Raphanus sativus*) and forage oats (*Avena sativa*).

| Camera | Camera ID | Sensor | Sensor type | Shutter type | Pixel area (μm²) | Image size (pixels) | OWL software | Non-default camera settings |
|---|---|---|---|---|---|---|---|---|
| Raspberry Pi v2 | V2 | Sony IMX219 | CMOS | Rolling | 1.25 | 416 × 320 | 29/05/2022 | Exp. mode: 'beach' Exp. comp.: -4 Sensor mode: 0 Res.: 416 × 320 |
| Raspberry Pi HQ | HQ1 | Sony IMX477 | CMOS | Rolling | 2.40 | 640 × 480 | 23/8/2021 | None |
| Raspberry Pi HQ | HQ2 | Sony IMX477 | CMOS | Rolling | 2.40 | 416 × 320 | 29/05/2022 | Exp. mode: 'beach' Exp. comp.: -4 Sensor mode: 0 Res.: 416 × 320 |
| Arducam | ARD | AR0234 | CMOS | Global | 9.00 | 416 × 320 | 29/05/2022 | Res.: 416 × 320 |

Targeting events e1.1 – 1.3 in the SSWC event timeline, field use and expected framerates were simulated by collecting video data for each camera while running the OWL detection software, following the method of Coleman et al. (2022b). The V2, HQ1 and HQ2 cameras work natively with the Raspberry Pi and OWL software using the library 'Picamera', whilst the ARD camera required the installation of a modified Raspberry Pi kernel driver and used the 'libcamera' library. The OWL detection software relies on green detection using the excess green (ExG) index (Woebbecke et al., 1995) combined with thresholding in the hue, saturation and value (HSV) colour space to reduce false detections in areas with bright reflections. The details of the algorithm are detailed in Coleman et al. (2022).

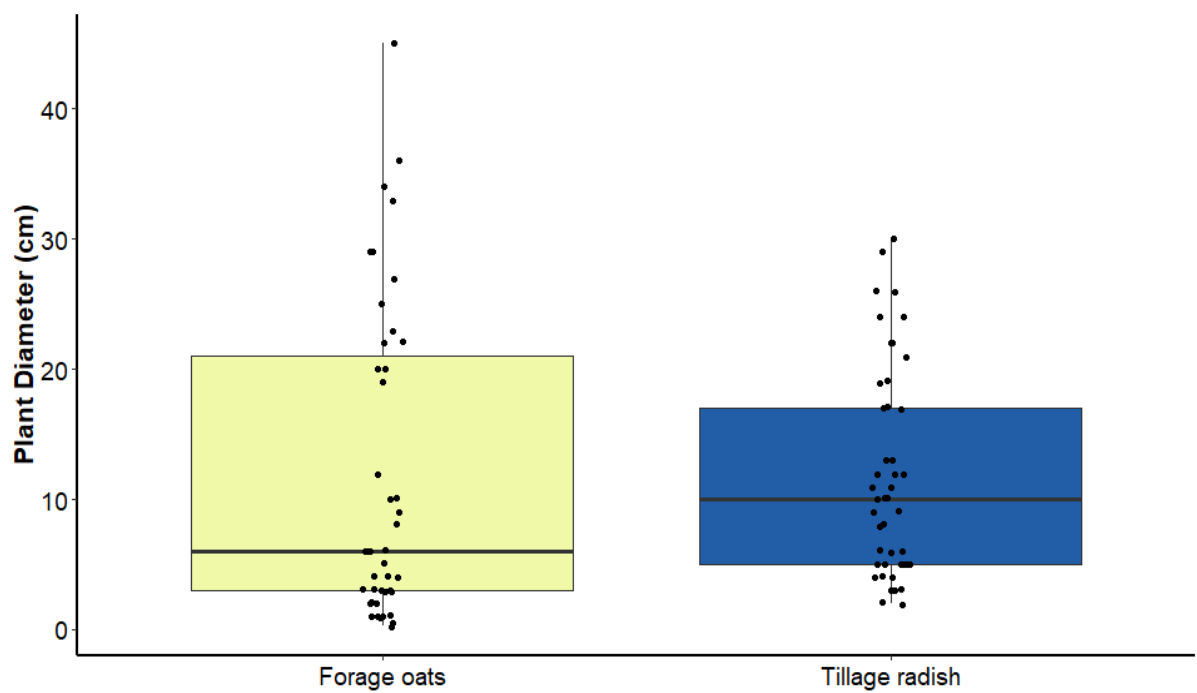

**Figure 3** Box and whisker plots of measured diameters of forage oat (n = 43) and tillage radish (n = 47) plants sampled within three randomly assigned 1 m² quadrats from each of the six transects.

*3.3    Data collection*

With the four-camera system, video data were collected across the six transects at five different speeds, 5, 10, 15, 20 and 30 km h$^{-1}$ in May 2022 (Figure 2). During each run, vehicle speed was maintained with the vehicle speedometer, whilst separately a rear-mounted GPS unit (ELEMNT ROAM; Wahoo Fitness, Atlanta, GA, USA) logged and checked true ground speed. Video footage from a GoPro Hero5 (GoPro Inc., San Mateo, California) was used as a high-resolution standard against which to compare detections. Camera lenses were checked for dust and contamination after each run. Technical issues with the Arducam AR0234 camera and the automatic gain adjustment resulted in only the first three transects being recorded correctly, this limitation is considered further in the Discussion.

Tillage radish and forage oats in each transect were counted manually prior to video data collection. The diameters of plants within three randomly allocated 1 m$^2$ quadrats in each transect were also measured prior to data collection (Figure 3).

*3.4    Analysing video data*

Following image data collection, video footage was analyzed on a frame-by-frame basis. Detections (red boxes in Figure 4) were compared with a high-resolution video, using custom Python-based software (accessible at: https://github.com/geezacoleman/OpenWeedLocator/blob/main/video_analysis.py), following the method of Coleman et al. (2022). If a detection was made but no plant was observed, then it was noted as a false positive (Figure 4). In some cases, where small plants other than tillage radish and forage oats had emerged in the field of view, the detection was classified as 'other' or 'moss'. This was not included in the false positive or true positive counts.

Based on these detection results, performance for each camera, at each speed was determined using the metrics of precision (Eq. 1) and recall (Eq. 2). The total number of weeds present was established from the manual plant counts of each transect prior to video collection. Additionally, recall was calculated on a per species basis. Given the detection algorithm does not include classes, species-wise calculation of precision was not possible.

$$\text{Precision} = \frac{True\ Positives}{True\ Positives + False\ Positives} \quad (1)$$

$$\text{Recall} = \frac{True\ Positives}{Total\ weeds} \quad (2)$$

*3.5    Blur assessment*

An estimate of non-referenced, image blur at e1.1 was calculated by analysing changes in the high-frequency components of Fast Fourier Transformed (FFT) images. The high frequency components of an image typically represent fine details such as edges that contribute to the overall clarity of an image. Whole-image motion blur from a moving vehicle often reduces these high frequency components and results in a lower value overall. The approach is based on a

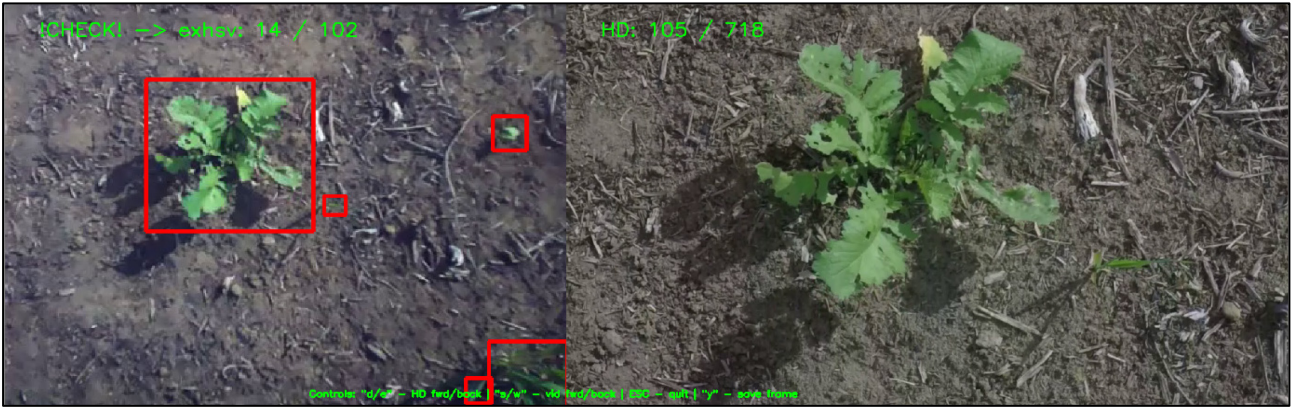

**Figure 4** The custom, frame-by-frame analysis tool developed in Python used to determine true and false positive weed detection rates for each video collected. Red squares indicate a detection. If needed, these were compared with a high-resolution video to determine if a weed is present in the field of view of the camera.

Python implementation using 'numpy' (Harris et al., 2020) and 'OpenCV' (Bradski, 2000) accessible at: https://github.com/geezacoleman/OpenWeedLocator/blob/0e17c7891f864573c9637bee60197cb41e886a69/utils/blur_algorithms.py.

*3.6 Statistical analysis*

Performance metrics (recall and precision) and blur data were analyzed in R Studio (RStudio Team, 2015). Fisher's protected Least Significance Difference (LSD; $\alpha = 0.05$) was used for pairwise comparisons of recall for the speed × camera interaction using the 'agricolae' package (de Mendiburu, 2020). Normality and homoscedasticity of the data were confirmed with Shapiro-Wilkes and Bartlett's tests, respectively. Differences in precision were compared using non-parametric Kruskal-Wallis test, followed by the Wilcoxon Rank Sum Test for pair-wise comparisons ($\alpha = 0.05$), using a Benjamini-Hochberg adjustment. Due to technical camera issues, ARD 20 km h$^{-1}$ had only two replicates and was removed from pairwise comparisons; however, it was included in regression analyses.

Two-factor (camera × class) regression analyses for speed and recall were performed in base R with the lm() function. Comparisons of performance between individual classes were conducted using Analysis of Covariance (ANCOVA) method, controlling for vehicle speed to observe underlying differences with species. Pairwise comparisons were made with the 'emmeans' package using the *Bonferroni* adjustment for cameras at each class level and classes at each camera level. Prior to analysis interactions between groups were checked with a Type II ANOVA. The normality of residuals was confirmed with the Shapiro test ($P > 0.05$) from the 'rstatix' package (Kassambara, 2023). Homogeneity of residuals was assessed with the Levene test ($P > 0.05$). Data were visualized and all figures were created using 'ggplot2' (Wickham, 2016).

## 4 Results

*4.1 Speed and performance*

ARD had the highest ($P < 0.05$) recall of all camera systems tested (Table 2), with up to 95.7% of weeds recalled at 5 km h$^{-1}$ and 85.7% at 30 km h$^{-1}$. Although there was an apparent decline in ARD performance with increasing speed, there were no differences (P < 0.05) in recall between the lowest and highest speeds (Figure 5). The lowest (P < 0.05) recall results were recorded for the HQ1 and V2 cameras at 30 km h$^{-1}$ and 15 – 30 km h$^{-1}$ speeds, respectively. Decreases in recall (P < 0.05) were observed when speed increased from 5 – 30 km h$^{-1}$ for HQ1, HQ2 and V2. The largest decrease of 25.7% was observed for HQ1. The same camera with updated software (HQ2) declined by a similar 24.3%, though performance was between 18.0 – 19.5% higher (P < 0.05) at all speeds.

**Table 2** Summary of camera performance at each of the five speeds tested. Letters indicating significant differences for recall using Fisher's protected least significance difference (LSD; $\alpha = 0.05$) are presented. The non-parametric pairwise Wilcoxon Rank Sum Test was performed to compare means for precision using a Benjamini-Hochberg adjustment ($\alpha = 0.05$). The top performing recall for each speed is indicated with red text.

[†] ARD at 20 km h$^{-1}$ was excluded from pairwise means comparisons due to its low sample size from technical issues with the camera.

| Camera | Speed (km h$^{-1}$) | Recall (%) ± SE | | | P < 0.05 | Precision (%) ± SE | | | P < 0.05 |
|---|---|---|---|---|---|---|---|---|---|
| ARD | 5 | **95.7** | **±** | **2.67** | *a* | 95.9 | ± | 1.53 | *b* |
| | 10 | **88.3** | **±** | **5.10** | *a* | 98.3 | ± | 0.23 | *b* |
| | 15 | **91.4** | **±** | **2.05** | *a* | 99.4 | ± | 0.62 | *ab* |
| | *20*[†] | ***84.6*** | **±** | ***0.07*** | - | *100* | ± | *0* | - |
| | 30 | **85.7** | **±** | **4.46** | *ab* | 99.6 | ± | 0.42 | *ab* |
| HQ2 | 5 | 74.8 | ± | 2.21 | *bc* | 100 | ± | 0 | *a* |
| | 10 | 67.1 | ± | 1.80 | *cd* | 100 | ± | 0 | *a* |
| | 15 | 60.5 | ± | 2.24 | *de* | 100 | ± | 0 | *a* |
| | 20 | 55.4 | ± | 2.28 | *ef* | 99.7 | ± | 0.31 | *ab* |
| | 30 | 50.5 | ± | 3.83 | *fg* | 100 | ± | 0 | *a* |
| HQ1 | 5 | 56.8 | ± | 2.45 | *ef* | 99.7 | ± | 0.27 | *ab* |
| | 10 | 48.9 | ± | 3.16 | *fg* | 100 | ± | 0 | *a* |
| | 15 | 45.9 | ± | 3.42 | *gh* | 100 | ± | 0 | *a* |
| | 20 | 39.3 | ± | 4.83 | *hij* | 100 | ± | 0 | *a* |
| | 30 | 31.1 | ± | 4.66 | *jkl* | 100 | ± | 0 | *a* |
| V2 | 5 | 45.7 | ± | 2.37 | *ghi* | 100 | ± | 0 | *a* |
| | 10 | 36.8 | ± | 3.34 | *ijk* | 100 | ± | 0 | *a* |
| | 15 | 32.9 | ± | 2.67 | *jkl* | 100 | ± | 0 | *a* |
| | 20 | 29.8 | ± | 3.52 | *kl* | 100 | ± | 0 | *a* |
| | 30 | 26.0 | ± | 4.04 | *l* | 100 | ± | 0 | *a* |

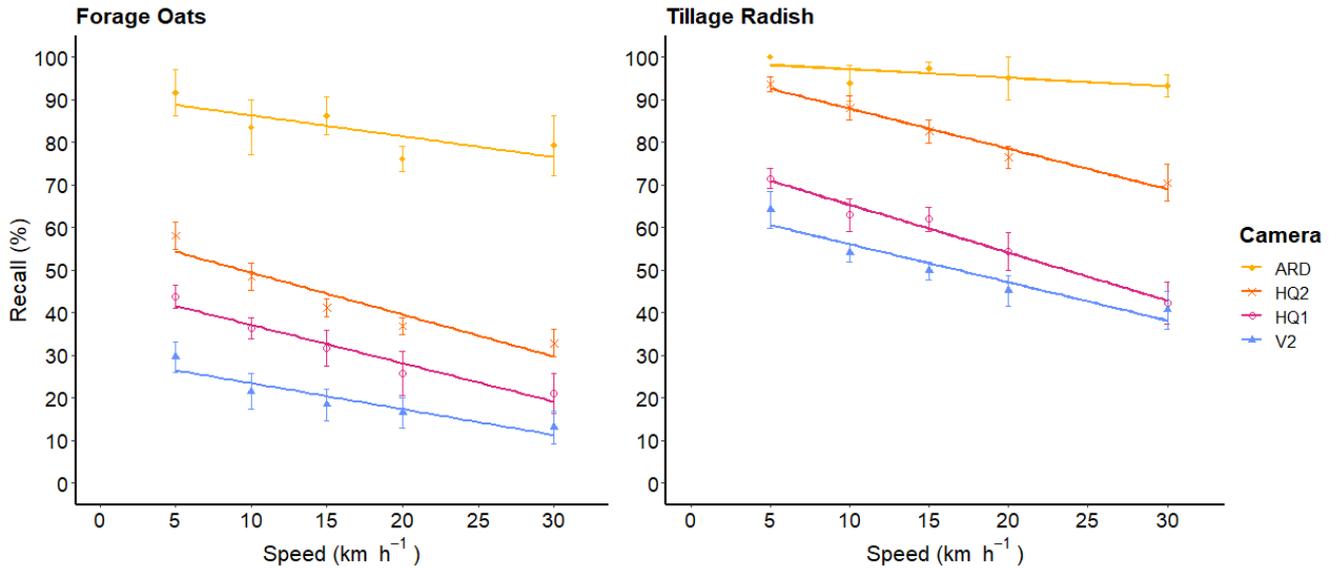

**Figure 5** Regression analysis for the influence of speed on recall by camera (ARD, HQ2, HQ1, and V2), and class; forage oats (*Avena sativa*; left) and tillage radish (*Raphanus sativus*; right). Differences were found for the slope (P < 0.05) and intercept (P < 0.05) for different camera-software combinations. Error bars with standard errors of the mean (n[HQ1, HQ2, V2]=6; n[ARD:5-15, 30km h$^{-1}$]=3; n[ARD:20km h$^{-1}$]=2) are included. Equation: $y = recall \sim speed * camera * class$.

HQ1, HQ2 and V2 had 100% precision for all speeds tested, dipping to 99.7% for HQ2 at 20 km h$^{-1}$ and HQ1 at 5 km h$^{-1}$. ARD recorded the lowest precision (P < 0.05), of 95.9% at 5 km h$^{-1}$ and 98.3% at 10 km h$^{-1}$.

The effect of increasing speed on recall for the ARD camera system was not linear for oats (P = 0.12) or tillage radish (P = 0.18), though a negative trend was apparent. There were negative linear relationships (P < 0.05) for speed and recall with the HQ1, HQ2 and V2 systems for both forage oats and tillage radish (Figure 5). The steepest slope was observed with the HQ1 camera system with 1.12% reduction in recall

**Table 3** Summary of regression parameters for the influence of speed on recall for camera by class interactions. Significant linear relationships (P<0.05) between speed and recall were found for HQ2, HQ1 and V2 camera systems for both tillage radish and forage oats. No linear relationship was observed for ARD. Equation: $y = recall \sim speed * camera * class$.

| Plant Species | Camera | Intercept (recall %) | Slope (Δ recall/km h$^{-1}$) | R$^2$ | P-value |
|---|---|---|---|---|---|
| Tillage radish (*Raphanus sativus*) | ARD | 99.20 | -0.20 | 0.50 | 0.18 |
| | HQ2 | 97.30 | -0.94 | 0.98 | < 0.01 |
| | HQ1 | 76.60 | -1.12 | 0.98 | < 0.01 |
| | V2 | 65.10 | -0.89 | 0.90 | 0.01 |
| Forage oats (*Avena sativa*) | ARD | 91.20 | -0.49 | 0.61 | 0.12 |
| | HQ2 | 59.30 | -0.98 | 0.89 | 0.02 |
| | HQ1 | 46.10 | -0.90 | 0.95 | < 0.01 |
| | V2 | 29.50 | -0.61 | 0.86 | 0.02 |

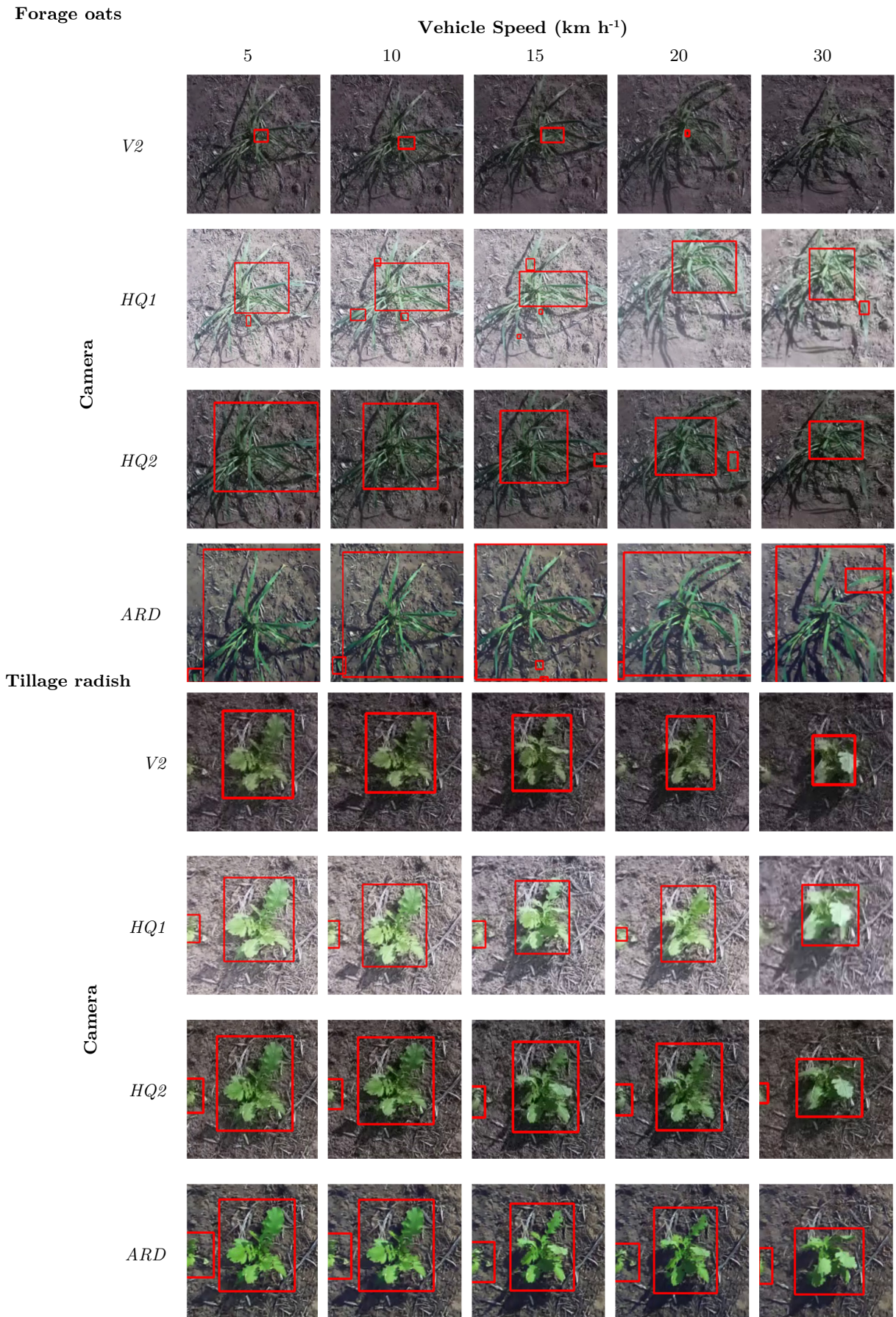

**Figure 6** Visualising the effect of vehicle speed and camera/software combination on detection performance for forage oats (top) and tillage radish (bottom). The same plant for forage oats and tillage radish was identified in each frame to allow for better comparison.

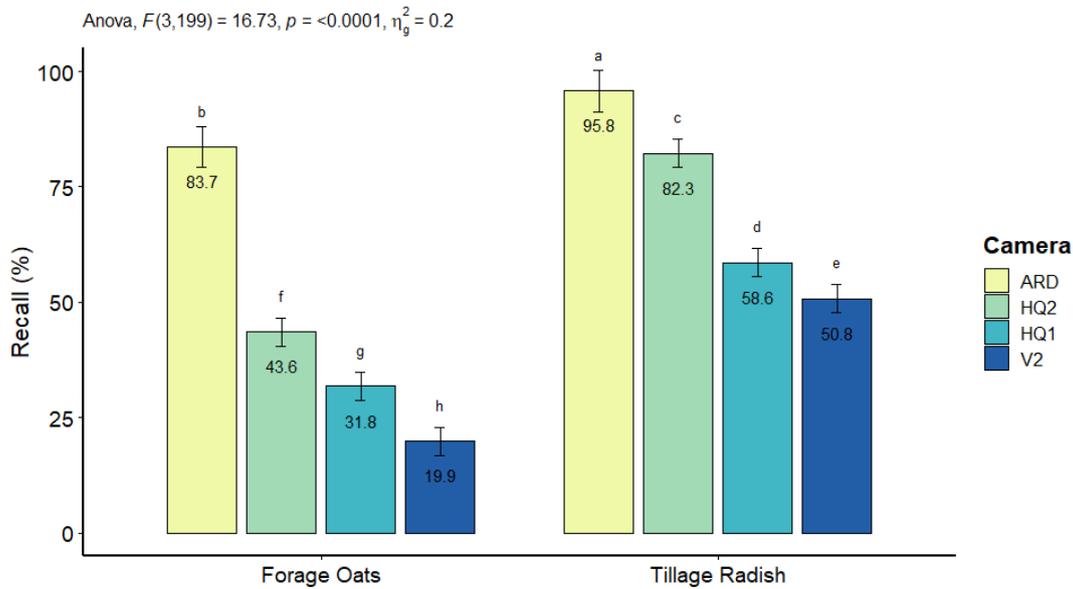

**Figure 7** Comparison of camera recall of forage oats (*Avena sativa*) and tillage radish (*Raphanus sativus*);. A two-way analysis of covariance (ANCOVA) with emmeans was used to control for speed to determine effects of plant class effects at a camera level. Controlling for speed all cameras were capable of higher recall on tillage radish than forage oats. Tests were conducted with Bonferroni adjustment and significance at P < 0.0125.

per km h$^{-1}$ for tillage radish and 0.90% reduction per km h$^{-1}$ for forage oats (Table 3).

A qualitative inspection of individual camera frames, (Figure 6) illustrates the change in image detection quality for forage oats and tillage radish over different speeds. The same plants for forage oats and tillage radish are observed in each frame shown. During e1.1, whole image motion blur appears to increase along with a greater wind effect from vehicle movement. This increases plant movement and dust around the camera, resulting in smaller detection sizes. With entirely default settings, the exposure of HQ1 is long and the images are overexposed.

*4.2    Influence of plant species on performance*

The broadleaved tillage radish was consistently detected more successfully (P < 0.05) than forage oats by all cameras (Figure 7). ARD consistently outperformed the other camera systems tested, with a mean recall of 95.8% for tillage radish and 83.7% for forage oats. Additionally, the 12.1% decrease in recall between tillage radish and forage oats by the ARD camera system was much less than the reductions in recall of 26.8%, 38.7% and 30.9% for HQ1, HQ2 and V2, respectively.

*4.3    Image blur*

The reported FFT blur value is a measure of the proportion of high frequencies components (e.g., edges and high contrast regions) within a Fourier-transformed image, whereby blur results in fewer high frequencies present. It separates algorithm performance at e1.2 from hardware effects at e1.1 (Figure 1). Speed had a comparatively reduced impact on FFT-measured blur for ARD camera system, with a non-significant (P = 0.055) relationship. The negative relationship between speed and FFT-based blur for HQ1, HQ2 and V2 camera systems confirms that speed resulted in increasingly blurry images (Figure 8). For the ARD camera system, there was a strong positive (R = 0.85; P < 0.01) linear relationship between image blur and recall of forage oats; however, there was not a similar relationship (P=0.083) for tillage radish (Figure 9), where recall of the broadleaf plant remained high. A significant positive correlation (R = 0.54; P < 0.01) was also found for HQ1 and tillage radish recall.

**5    Discussion**

The ARD camera system outperformed the other three camera systems tested in maintaining high weed detection performance across all working speeds used in this evaluation. The recall of ARD of 95.7% of weeds at 5 km h$^{-1}$ and 85.7% at 30 km h$^{-1}$ was at least 20% higher than those of the other camera systems (Table 2). Importantly, the high recall was achieved whilst maintaining precision above 95.9%, up to 99.6% at 30 km h$^{-1}$. The increase in precision with speed, suggests the ARD is more sensitive to false positives at low speeds. The lower native resolution but global shutter ARD has the clearest images (Figure 6), with a substantially larger pixel area of 9 µm$^2$ compared to the other camera hardware tested. The larger pixel area acts as a greater 'catchment area' for photons, increasing the signal to the sensor. Thus, it can often improve the dynamic range of the camera, with each pixel capable of storing more charge, reducing clipping across brightness levels. Colour accuracy is also improved with larger pixel sizes resulting in a higher signal-to-noise ratio. These factors are likely a contributing factor to the high performance of ARD over the other systems tested.

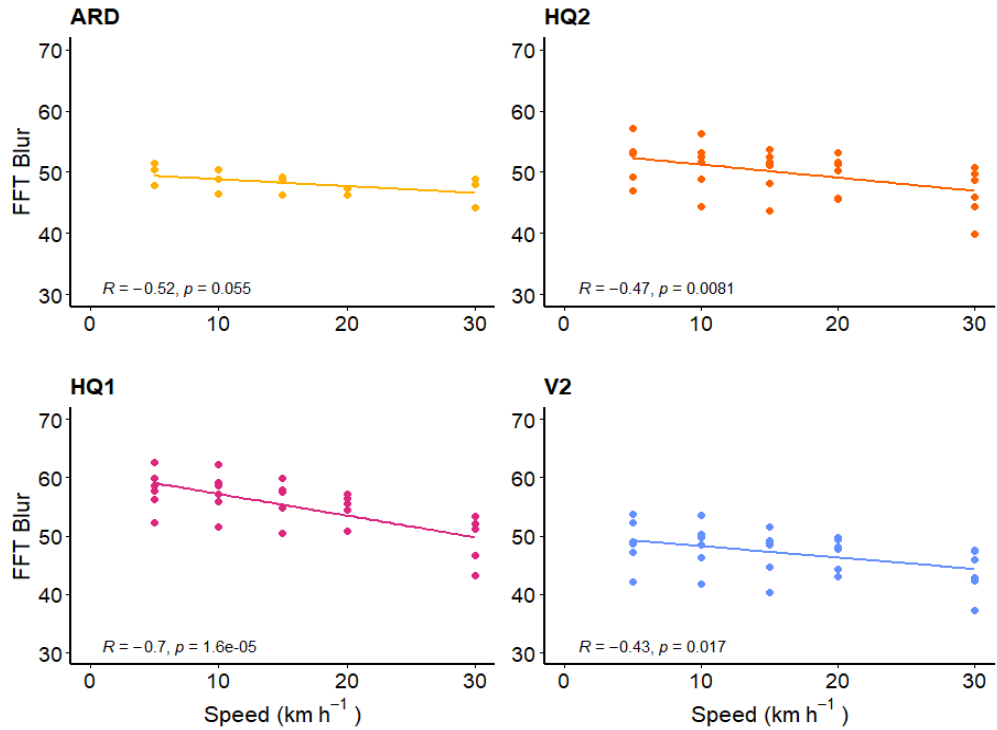

**Figure 8** Using fast-Fourier transform (FFT)-based analysis, blur reduces the proportion of high frequency components (e.g., edges) within an image, thus a higher FFT blur value indicates a less blurry image. There was no significant (P = 0.055) correlation between speed and based blur analysis for ARD. Negative correlations (P < 0.05) were observed for HQ2, HQ1 and V2 camera systems, indicating speed resulted in increased blur. Pearson's correlation coefficient was used for analysis.

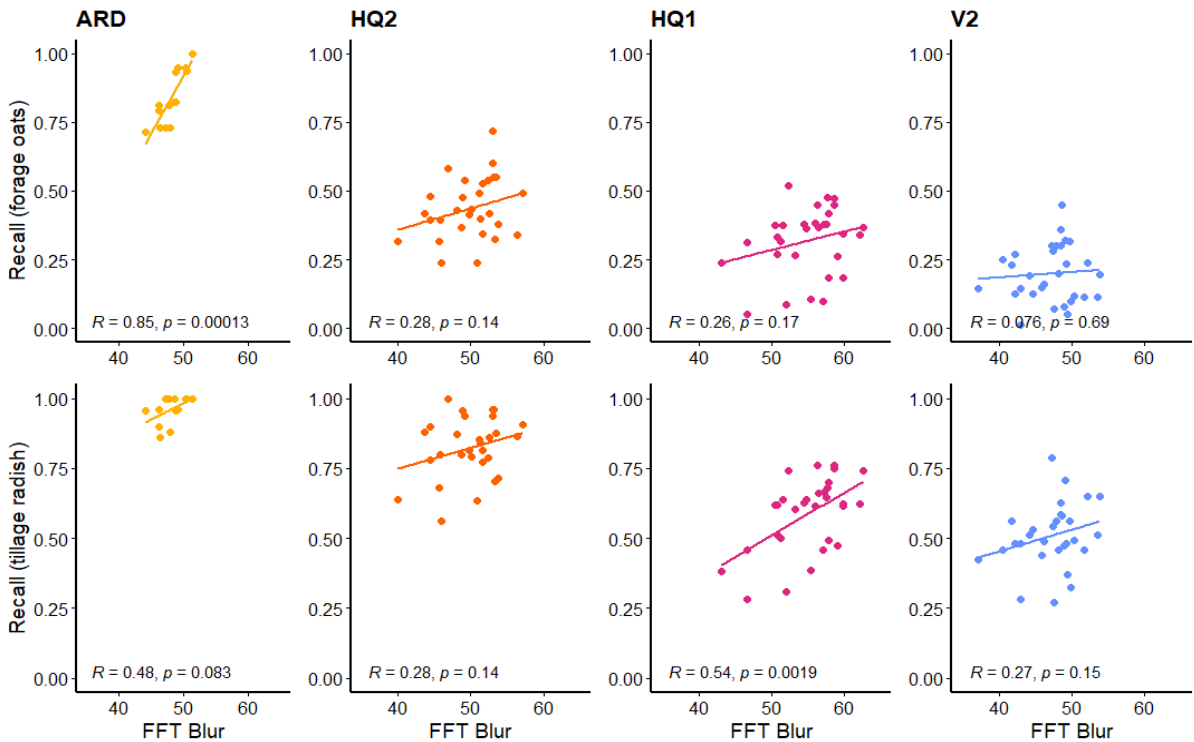

**Figure 9** Investigating the relationship between image blur, as measured using Fast-Fourier Transform (FFT)-based frequency analysis, and recall for forage oats and tillage radish. Significant positive correlations were observed for ARD and forage oat recall, and HQ1 and tillage radish recall. Higher FTT blur values indicate less blurry images, thus a positive relationship suggests blur is impacting detection performance, however, existing low performance (in the case of HQ1 and V2) would obscure correlations, where performance is unlikely to decrease further.

Whilst similar research targeting the combined e1.1 – 1.3 stages is limited, the 1.11 % decrease in recall reported in Steward et al. (2002) per km h$^{-1}$ increase in speed is substantially greater than a suggested 0.4% decrease in recall per km h$^{-1}$ for ARD between 5 – 30 km h$^{-1}$ found in this study. The decline here, is also considerably lower than that of Liu et al. (2021), who reported a decrease by 2.25% decrease per km h$^{-1}$ when increasing speed up to 5km h$^{-1}$, even with the use of more advanced deep learning, image classification algorithms. In that study, the authors used low cost 'webcams' (Aluratek AWC01F) and found that image blur was substantial above 3 km h$^{-1}$. Whilst image augmentation was included in the training pipeline, the training dataset was collected while stationary with high quality cameras, limiting model exposure to blur and likely reducing generalizability (Hu et al., 2021).

The performance of HQ1 and HQ2 are consistent with the results reported in our previous work (Coleman et al., 2022b), whilst the maximum speed tested was increased from 4km h$^{-1}$ up to 30 km h$^{-1}$, a more realistic range of speeds for large-scale weed control systems (Butts et al., 2021). Comparing the default settings on the Raspberry Pi HQ camera (HQ1) with the modifications made to the camera settings and image processing code in the OWL repository (HQ2), the HQ2 camera consistently outperformed HQ1, even with the lower resolution. Software changes between HQ1 and HQ2 were made to improve the efficiency of the system after identifying unnecessarily high exposure levels in default settings and inefficient methods of post-detection management. The higher resolution of the HQ1 camera resulted in fewer frames collected per transect and the missing of key weeds not just by the detection algorithm but by complete lack of capture in the recorded videos, particularly at faster travel speeds. The higher default exposure/brightness levels of the HQ1 camera are also more adversely impacted by blur, as the camera prioritises higher brightness images by lengthening exposure times. The HQ2 is now the current standard camera and software with the OWL system. Whilst the V2 camera has the same software-based settings as the HQ2, the image is of a lower quality, likely linked to the smaller sensor size (1.25 µm$^2$ vs 2.40 µm$^2$ pixel size respectively). With many machine vision applications using down-sampled images from the native resolution of the camera to reduce processing time (Hasan et al., 2021), pixel size, rather than native sensor resolution is likely the most important factor to improving weed detection performance, given the high performance of ARD.

Importantly, the results here underscore the strong and differential impact of speed on the image-based detection of two species with different growth habits, namely *broadleaved* tillage radish and the '*grassy*' forage oats. Controlling for speed, there was a 38.7% difference in performance for HQ2 between the broadleaf tillage radish and the graminaceous forage oats. Observing individual frames from videos in Figure 6, the thinner leaves of forage oats are impacted more by blur and plant movement than the broad, lobed leaves of the tillage radish. This is confirmed by the strong correlation between forage oats recall and image blur for the ARD camera. With differential performance based on plant species, there is a risk that such vision-based systems may unintentionally alter the in-field distributions of weed species, by preferentially removing one over others. Increasing speed of the weed detection device may result in uneven and unexpected declines in performance. Combined with detection size analysis, it appears that small detections will be most likely missed at higher speeds, with larger weeds continuing to be found based on larger central green masses.

## 6 Conclusion

Our study has shown that it is possible to reliably detect weeds in fallow at speeds of up to 30 km h$^{-1}$ using colour based algorithms and relatively low-cost hardware. The negative relationship between speed and weed detection performance for three of the four camera systems tested in this study highlights that camera system selection should be a top priority in the development of new SSWC systems. The impact of camera system and speed on detection performance is amplified when we consider the differences in detection performance for broadleaf (more easily detected, less affected by speed) vs. grass weeds (less easily detected, more affected by speed). It is likely that larger pixel area will provide advantages over cameras with smaller pixel areas, even if the overall sensor has a lower native resolution, due to benefits in dynamic range and signal-to-noise ratio for colour fidelity. Whilst the results presented indicate that digital camera systems can now perform at the speeds required by industry, it is important to remember that weed detection by the camera system is the first step in a sequence of events required to achieve appropriate weed control with camera based SSWC systems (outlined in Figure 1). The next step in understanding how to reliably achieve appropriate real time weed control performance is investigation of complete system performance, from the weed entering the camera field of view to completion of weed control, under a variety of conditions (e.g., day, night, sunny, overcast). For operation at the speeds tested here, this will require fast signaling from computer systems, fast actuation of solenoids and fast delivery of herbicide from spray nozzle to the weed.

## 7 Declaration of Competing Interest

The authors declare that they do not have any commercial or associative interest that represents a conflict of interest in connection with the work submitted.

## 8 Acknowledgements

This work was supported by the Grains Research and Development Corporation grant *Innovative crop weed control for northern region cropping systems* (US00084).## 9 References